\documentclass[]{interact}

\usepackage{epstopdf}% To incorporate .eps illustrations using PDFLaTeX, etc.
\usepackage[caption=false]{subfig}% Support for small, `sub' figures and tables
\usepackage[natbibapa,nodoi]{apacite}% Citation support using apacite.sty. Commands using natbib.sty MUST be deactivated first!
\theoremstyle{plain}% Theorem-like structures provided by amsthm.sty

\theoremstyle{definition}

\theoremstyle{remark}

\usepackage{graphicx}
\usepackage{booktabs}
\usepackage{stfloats}
\usepackage{multirow}
\usepackage{amsmath}
\begin{document}

\articletype{RESEARCH ARTICLE}% Specify the article type or omit as appropriate

\title{Joint Learning of Wording and Formatting for Singable Melody-to-Lyric Generation}

\author{
\name{Longshen Ou, Xichu Ma, and Ye Wang\textsuperscript{a}\thanks{CONTACT Ye Wang Email: 
wangye@comp.nus.edu.sg}}
\affil{School of Computing, National University of Singapore, 21 Lower Kent Ridge Road, Singapore}
}

% \author{
% \name{A.~N. Author\textsuperscript{a}\thanks{CONTACT A.~N. Author. Email: latex.helpdesk@tandf.co.uk} and John Smith\textsuperscript{b}}
% \affil{\textsuperscript{a}Taylor \& Francis, 4 Park Square, Milton Park, Abingdon, UK; \textsuperscript{b}Institut f\"{u}r Informatik, Albert-Ludwigs-Universit\"{a}t, Freiburg, Germany}
% }

% \author{
% \name{Longshen Ou, Xichu Ma, and Ye Wang\textsuperscript{a}\thanks{CONTACT A.~N. Author. Email: latex.helpdesk@tandf.co.uk} and John Smith\textsuperscript{b}}
% \affil{\textsuperscript{a}Taylor \& Francis, 4 Park Square, Milton Park, Abingdon, UK; \textsuperscript{b}Institut f\"{u}r Informatik, Albert-Ludwigs-Universit\"{a}t, Freiburg, Germany}
% }

\maketitle

\begin{abstract}

Despite progress in melody-to-lyric generation, a substantial singability gap remains between machine-generated lyrics and those written by human lyricists. In this work, we aim to narrow this gap by jointly learning both wording and formatting for melody-to-lyric generation. After general-domain pretraining, our model acquires length awareness through an self-supervised stage trained on a large text-only lyric corpus. During supervised melody-to-lyric training, we introduce multiple auxiliary supervision objective informed by musicological findings on melody--lyric relationships, encouraging the model to capture fine-grained prosodic and structural patterns. Compared with naïve fine-tuning, our approach improves adherence to line-count and syllable-count requirements by 3.8\% and 21.4\% absolute, respectively, without degrading text quality. In human evaluation, it achieves 42.2\% and 74.2\% relative gains in overall quality over two task-specific baselines, underscoring the importance of formatting-aware training for generating singable lyrics.\footnote{Code and data will be released upon acceptance.}

\end{abstract}

% \footnote{Code available at \url{https://github.com/Sonata165/BART-M2L}}.


\section{Introduction}\label{sec:introduction}

% Move 1: research territory, important, central, interesting, problematic, relevant
Natural language processing has become increasingly intertwined with music-related applications \citep{doi:10.1080/09298215.2018.1488878}. Among them, automatic lyric generation has drawn growing attention from both research and industry. A particularly important task is \emph{melody-to-lyric generation} (M2L), which aims to produce lyrics that can be naturally sung with a given melody. A reliable M2L system would substantially reduce the cost and effort involved in songwriting and music production \citep{liu2022chipsong}.

\begin{figure}[tb]
\centering
\resizebox*{\columnwidth}{!}{
% \pdftooltip{\includegraphics{resources/intro_annot_flatten_crop.pdf}}{This is a alt text of the lyric.}
\includegraphics{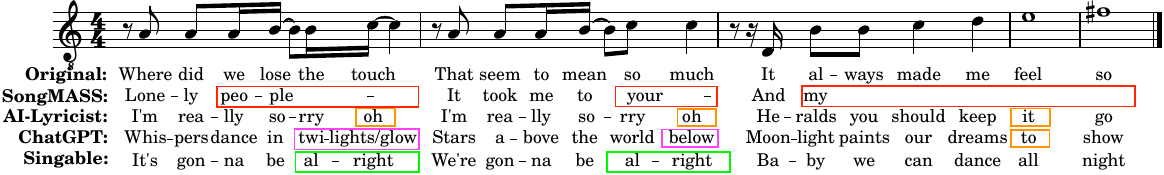}
}
\caption{\textbf{Caption}: A comparative illustration of original human-written lyrics and the outputs produced by SongMASS, AI-Lyricist, ChatGPT, and our proposed system, all conditioned on the same melody. Highlighted regions indicate syllable--note mismatches or incorrect prosodic alignment in baseline models, contrasted with the improved alignment in our system. \\\hspace{\textwidth} \textbf{Figure 1. Alt Text}: The figure displays a musical staff with a short melody, followed by five rows of lyrics: the original human-written lyrics, and the outputs of SongMASS, AI-Lyricist, ChatGPT, and our proposed model. Colored boxes highlight issues such as incorrect syllable–note alignment, unnatural stress placement, and mismatched vowel durations in the baseline systems. In contrast, the lyrics generated by our model show correct syllable counts and more natural alignment with melodic peaks and note durations.} 
\label{fig:intro}
\end{figure}

% More background
Lyrics are crafted to be sung. Their \emph{singability}---the degree to which they can be comfortably sung with the intended melody---is therefore central. Although definitions vary, we follow the perspective of \citet{low2003singable}, viewing singability as shaped by both linguistic and musical factors. Linguistically, lyrics should use pronounceable language and avoid overly complex consonant clusters or extremely short vowels. Musically, the wording should align with melodic phrasing and note durations so that the pronunciation of lyrics do not conflict with melody expression. With modern pretrained language models, adjusting the linguistic style can be handled with modest fine-tuning. In contrast, modeling music–lyric compatibility is considerably more challenging. This work focuses specifically on improving this compatibility in M2L generation, without adding constraints such as keywords or emotional control.

% Move 2: Niche: gap, counter-claiming, raising the question, continuing a tradition
If the melody–lyric relationship is not properly handled, lyrics that appear fluent on paper may sound awkward when sung. Figure~\ref{fig:intro} illustrates this issue using several phrases from \emph{Free as a Bird}, comparing the original lyric with outputs from SongMASS \citep{sheng2021songmass}, AI-Lyricist \citep{ma2021ai}, and ChatGPT \citep{openai2023gpt4}. SongMASS produces grammatically inconsistent text. AI-Lyricist and GPT-4 produce more fluent sentences, yet they fail to map the words appropriately to the musical structure: certain words must be stretched unnaturally (red boxes), multiple notes must be splitted to fit overcrowded syllables (purple boxes), and unimportant words are mistakenly emphasized due to misalignment with long notes (orange boxes). All these issues distort the intended rhythmic and expressive qualities of the melody and reduce singability.

% Niche continue
In contrast, the final example in Fig.~\ref{fig:intro} exhibits both linguistic fluency and appropriate melodic alignment. Its syllable count matches the number of notes, and stressed syllables fall on rhythmically prominent notes—for instance, the second syllable of \emph{alright} aligns with the longest note in the corresponding measure in the melody. Such alignment allows the lyric to be sung naturally and preserves the composer's intended rhythm.

% Occupying the niche, purpose and contribution
To address the challenges above, this paper presents a method for generating singable lyrics from melodic input. Our contributions are threefold:

\begin{itemize}
    \item We introduce a self-supervised training method that enables a paragraph-level text generation model to acquire robust \emph{length awareness}, controlling both the number of lines and the number of syllables per line—two fundamental requirements for lyric singability. This capability supports reliable melody–lyric alignment during inference and does not require paired melody–lyric data.
    \item We design several auxiliary supervised training objectives that encourages the model to learn melody-related textual patterns essential for singability. This strategy made better use of the limited paired data available for supervised M2L training, and improves fine-grained compatibility.
    \item By combining both self-supervised and supervised components, we develop a lyric generation system that achieves competitive overall generation quality and outperforms prior approaches in both quantitative and qualitative evaluations.
\end{itemize}

The rest of this paper is organized as follows. Section~2 reviews prior work on melody–lyric relationships and existing approaches to M2L. Section~3 describes the proposed method. Section~4 outlines datasets, baselines, training configurations, and evaluation metrics. Section~5 presents experimental results, including comparisons with external baselines and an ablation study. Section~6 concludes the paper with a summary of our explorations.

\section{Related Work}

This section reviews prior studies on melody--lyric relationships in human-composed songs, computational models that attempt to capture these relationships, and recent NLP approaches relevant to the M2L task.

\subsection{Relationship between melody and lyrics}

Unlike text-only lyric generation, melody-to-lyric (M2L) systems operate under explicit and implicit structural constraints imposed by the melody. These constraints influence the linguistic form, prosody, and syntactic shape of the generated text. While lyricists may flexibly interpret these constraints, most human-written lyrics still conform to them. Below, we summarize key factors identified in musicological and computational studies.

\paragraph*{Length and Syllable Alignment.}
A fundamental constraint arises from the number of notes in each melodic line, which determines the required number of syllables in the corresponding lyric. Exceeding this limit forces note subdivision, while falling short leads to syllable extension—both detrimental to rhythmic naturalness \citep{noske1988french}. Consequently, many M2L systems explicitly condition the generation process on the number of syllables per line \citep{lee2019icomposer, li2020rigid, ma2021ai, liu2022chipsong, guo2022automatic}.

\paragraph*{Stress and Rhythmic Compatibility.}
For stress-timed languages such as English, the placement of stressed syllables plays a crucial role in singability \citep{low2003singable}. Lyricists often align stressed syllables with metrically strong or salient notes. Several studies therefore attempt to control the position of stressed syllables in generated lyrics \citep{ghazvininejad2018neural, xue2021deeprapper}.

\paragraph*{Fine-Grained Melody--Lyric Correlations.}
\label{sec:related_singability}
Beyond global length and stress, more nuanced dependencies exist. \citet{nichols2009relationships} report that:
(1) stressed syllables correlate with metrical position, melodic peaks, and note duration;  
(2) stopwords are associated with metric position and melodic peaks;  
(3) vowel duration is related to note length.  
These observations reveal a multi-dimensional relationship between melodic and linguistic structure.

Additional melody--text interactions have been documented, including tone–pitch alignment in tonal languages \citep{Zhang:2021, guo-etal-2022-automatic} and genre-dependent vowel distribution patterns in pop music \citep{doi:10.1080/09298215.2021.1936076}. In this work we focus on English lyrics and do not explore genre- or tone-specific effects.

\paragraph*{Data Sparsity and Loose Coupling.}
An important characteristic of melody--lyric pairs is that their relationship is \emph{loosely coupled}: many melodies can fit the same lyrics, and a single melody may admit numerous plausible lyric realizations. This weak dependency leads to significant data sparsity, making it difficult for sequence-to-sequence models to learn such relationships from limited paired data. This motivates the use of stronger auxiliary supervision signal to guide learning.

\subsection{Melody-to-Lyric Generation}

\paragraph*{Rule-Based Approaches.}
Early M2L systems included rule-based approaches that constrained vocabulary choices based on prosodic compatibility with the melody \citep{oliveira2007tra, oliveira2015tra}. These systems could accurately satisfy length and stress constraints but often produced rigid lyrics that lacks fluency or naturalness.

\paragraph*{Supervised Sequence Modeling.}
Many recent works adopt a supervised, melody-conditioned text generation approach. Examples include melody-conditioned language models \citep{watanabe2018melody}, joint M2L and L2M modeling with LSTMs \citep{lee2019icomposer}, SeqGAN-based frameworks with additional constraints \citep{chen2020melody, ma2021ai}, unified dual-direction models with masked pretraining \citep{sheng2021songmass}, reconstruction-enhanced models \citep{qian2022training}, and approaches incorporating rhythm and tempo features \citep{li2022fuzzy}.  
\citet{zhang2024controllable} further propose syllable-level decoding and an explicit n-gram loss to enhance generation diversity.

\paragraph*{Unsupervised M2L.}
Due to the scarcity of paired data, recent work explores unsupervised/self-supervised M2L training using lyric-only corpora \citep{tian-etal-2023-unsupervised, qian-etal-2023-unilg}. These methods typically project the melody into a \textit{rhythm pattern}—a reduced representation capturing syllable counts—and use it as a structural condition. This line of work underscores the centrality of length control and demonstrates that length awareness can be learned without paired data.

\paragraph*{Remaining Challenges.}
Despite progress, several gaps remain. Many models still struggle to achieve singability, often misaligning notes and syllables \citep{ma2021ai, sheng2021songmass}. Supervised models theoretically could learn fine-grained melody--lyric correspondences, but limited paired data hampers this. Self-supervised approaches capture length but can overlook nuanced compatibility features such as stress dynamics, word importance, or vowel duration.

\subsection{Generate Lyrics with Other Inputs}

Beyond M2L, lyric generation has been studied under a variety of input conditions: length \citep{wu2019hierarchical, li2020rigid, liu2022chipsong}, stress patterns \citep{barbieri2012markov, xue2021deeprapper}, rhymes \citep{barbieri2012markov, Lingan2021AMD, zhang2022youling}, keywords \citep{nikolov2020rapformer}, emotional cues \citep{huang2021automated}, style \citep{Lingan2021AMD, chang2021singability}, structural templates \citep{lu2019syllable}, external text \citep{zhang2022qiuniu}, or musical accompaniment \citep{melistas2021lyrics, watanabe2021atypical}. Other works target specialized outputs such as structurally tagged songs \citep{potash2015ghostwriter} or lyrics containing hidden messages \citep{tong2019text}.

However, these techniques typically address only a subset of the fine-grained compatibility properties required for singable lyrics. In particular, factors such as word importance and vowel duration are rarely modeled. Moreover, existing stress control mechanisms require users to manually plan stress syllable position beforehand, which may not match realistic English stress distributions or the melody’s implicit requirements. Such planning is in fact particularly non-trivial. Even within human-written songs, not all long notes correspond to stressed syllables (only 82.45\% in our corpus; see \S\ref{sec:res_obj}), illustrating the need for flexible, data-driven modeling.

\subsection{Sequence-to-Sequence Denoising Pretraining}

Given the limited availability of melody--lyric parallel data, a strong pretrained model is crucial for maintaining fluency and semantic coherence. We therefore adopt denoising sequence-to-sequence pretraining \citep{lewis2020bart}, which has proven effective in tasks such as summarization \citep{akiyama-etal-2021-hie} and translation \citep{liu2020multilingual, tang2020multilingual, ou-etal-2023-songs}. Transfer learning from general-domain pretrained models has also consistently shown benefits in data-scarce settings \citep{gu2022mm, DBLP:conf/ismir/OuGW22}, making this approach well suited for M2L.

\subsection{Prompt-Based Fine-Tuning}

Prompt-based methods has become an influential paradigm in NLP \citep{liu2023pre}, where desired attributes of the target output are encoded as additional input to a pre-trained language model without modifications to the original objective function. Prompt-based fine-tuning has been shown to guide output properties such as lexicon selection \citep{susanto2020lexically, chousa-morishita-2021-input, wang-etal-2022-integrating}, content structure \citep{liu2021refsum}, output length \citep{lakew2019controlling}, or initial tokens \citep{li2022prompt}. In lyric generation, prompts have been employed to control syllable counts, stress patterns, and rhymes \citep{li2020rigid, ma2021ai, xue2021deeprapper, ormazabal2022poelm, liu2022chipsong}.  

\section{Methodology}

We aim to improve melody–lyric compatibility and thereby narrow the singability gap observed in existing systems. Our approach begins with paragraph-level length control using prompt-based methods, followed by additional training objectives that encourage aligning syllable stress, word importance, and vowel duration with their corresponding melody attributes.

\subsection{Problem Definition}

\begin{figure}[tb]
\centering
\resizebox*{\columnwidth}{!}{
\includegraphics{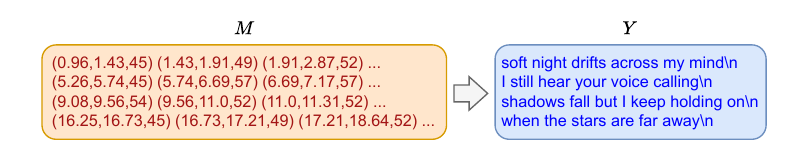}
}
\caption{\textbf{Caption}: An illustration of the paragraph-level melody-to-lyric generation task. The input melody \(M\) consists of multiple phrases, each represented as a sequence of note events encoded by onset, offset, and pitch. Given this structured melodic input, the system produces a lyric paragraph \(Y\) with the same number of lines, ensuring fluency, one-to-one alignment between notes and syllables, and more singability requirements. \\\hspace{\textwidth} \textbf{Figure 2. Alt Text}: A diagram illustrating the melody-to-lyric generation task. On the left, a rounded box labeled M contains a four sequences of note events. Each sequence correspond to one melody phrase, and each note event is represented as a triplet of onset time, offset time, and pitch (e.g., “(0.96, 1.43, 45)”). On the right, a rounded box labeled Y contains the generated four-line lyric paragraph: “soft night drifts across my mind”, “I still hear your voice calling”, “shadows fall but I keep holding on”, and “when the stars are far away”. An arrow between the two boxes indicates generating lyrics from the input melody.} 
\label{fig:inout_eg}
\end{figure}

We formally define the melody-to-lyric (M2L) generation task at the \emph{paragraph level}, where a paragraph denotes a structural section of a song, such as a verse, chorus, or bridge. Figure~\ref{fig:inout_eg} illustrates an example of the input melody and its corresponding lyric output. Given the melody of a perticular music section, the model generates a multi-line lyric paragraph that is aligned with the melody both structurally and prosodically.

\paragraph*{Melody Representation.}
A \textit{melody paragraph} is represented as a sequence of $L$ \emph{phrases}, and each melody phrase are aligned with one lyric utterance/line. 
\[
M = [m^{(1)}, m^{(2)}, \dots, m^{(L)}].
\]
where each $m$ is an ordered sequence of $N$ note events:
\[
m = \{(t_i^{\text{on}},\; t_i^{\text{off}},\; p_i)\}_{i=1}^{N},
\]
where $t_i^{\text{on}}$ and $t_i^{\text{off}}$ are onset and offset times in seconds, and $p_i$ is the pitch rounded to the nearest semitone.

\paragraph*{Lyric Output.}
The model outputs a \textit{lyric paragraph}
\[
Y = [y^{(1)}, y^{(2)}, \dots, y^{(L)}],
\]
where each $y^{(\ell)}$ is a lyric line consisting of a sequence of words. In performance, each word is realized as one or more syllables, depending on the word's pronunciation. We assume strict one-to-one alignment between notes and syllables. 
Let \(T_\ell\) denotes the number of syllables in the $\ell$-th line.  
The alignment constraint is
\[
T_\ell = |m^{(\ell)}| \qquad \forall \ell,
\]
meaning the number of lyric syllables in each line must exactly equal the number of melody notes in the corresponding phrase.

\paragraph*{Constraints.}
The M2L task imposes several constraints:

\begin{itemize}
    \item \textbf{Textual fluency and grammaticality.}  
    Each lyric line must form a coherent, well-formed English utterance.

    \item \textbf{Length constraints.}  
    The number of lyric lines must match the number of melody phrases, and the syllable count in each line must match the number of melody notes in the corresponding phrase.

    \item \textbf{Fine-grained melody--lyric compatibility.}  
    Beyond length, the lyrics should align prosodically with the melody, reflecting natural stress patterns, vowel placement, and other aspects of human-created, singable lyrics.
\end{itemize}

\paragraph*{Objective.}
Given a melody paragraph $M$, the goal is to generate a lyric paragraph $Y$ that (1) satisfies length constraints, (2) maintains textual fluency, and (3) achieves high melody–lyric compatibility.

\subsection{Gaining Length Awareness in Self-Supervised Learning}

Achieving compatibility between lyrics and melody begins with synchronizing their lengths. Our approach aims to ensure every generated line meets its target syllable count exactly. This follows established practices in format-controlled text generation \citep{ou-etal-2023-songs, zhang2024controllable, ma2021ai, liu2022chipsong}. Such strict length control provides a reliable alignment between lyric syllables and musical notes, forming the foundation upon which finer-grained compatibility mechanisms can be effectively applied.

\subsubsection{Prompt-Based Control}
% Overall strategy, Prompt construction
To enforce length constraints during lyric generation, we adopt a prompt-based fine-tuning strategy. For each paragraph, we construct a sequence of length-control tokens, where each token \texttt{<len\_n>} specifies the required number of syllables for a particular line to $n$. The number of length tokens thus matches the number of lyric lines to be generated. Using the CMU Pronunciation Dictionary \citep{cmudict}, we obtain word–syllable mappings.\footnote{For out-of-vocabulary words, a rule-based syllable estimator is applied; see the code for details.} An additional token, \texttt{<b>}, is used to mark sentence boundaries. During training, the length tokens are integrated as supplementary input to enhance the model's awareness of length constraints.

% Training strategy
We adopt a foundation model that is already pre-trained with general domain text corpus, and then continue training with lyric data with these prompts to strengthen its length awareness. Given BART’s \citep{lewis2020bart} strong performance in sequence-to-sequence generation, we adopt it as our foundation model. We extend BART’s tokenizer and embedding matrix to accommodate the newly introduced length tokens, ensuring each prompt token is represented as a learnable vector in the model’s hidden space.

\subsubsection{Length-Aware Self-Supervised Training}

% This method should work well if we have a large amount of data. 
Learning reliable length control requires substantial data. As shown in our ablation study (\S\ref{sec:ablation}), the limited amount of paired melody--lyric data is insufficient for a model to acquire robust length awareness on its own. To address this, we leverage large-scale text-only lyric corpora, which are far more abundant than paired datasets. We introduce an additional training stage—conducted before supervised M2L fine-tuning—dedicated to teaching the model how to follow the length requirements.

% , and meanwhile, to adapt its output style to the lyric domain.

\begin{figure}[tb]
\begin{center}
\includegraphics[]{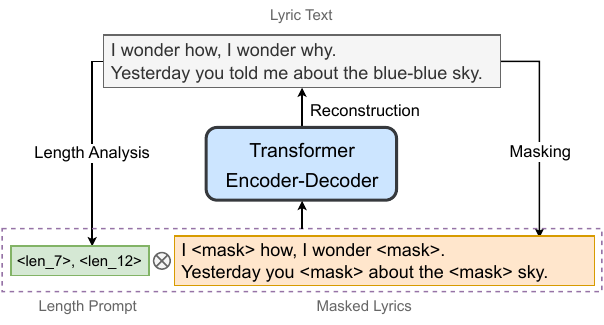}
\caption{\textbf{Caption}: Illustration of the self-supervised length-aware training process. 
Sentence-level length prompts (derived from syllable counts) are concatenated with 
masked lyric text and fed into a Transformer encoder--decoder model, which is trained 
to reconstruct the original lyrics using both the corrupted input and the provided 
length constraints. The symbol $\otimes$ denotes sequence concatenation. 
\\\hspace{\textwidth} \textbf{Figure 3. Alt Text}: 
A diagram showing the self-supervised training pipeline for 
length-aware learning. The original two-line lyric paragraph (“I wonder how, I wonder why.” 
and “Yesterday you told me about the blue-blue sky.”) is processed in two steps: 
(1) a length analysis module determines the syllable count of each line, producing 
length tokens such as \texttt{<len\_7>} and \texttt{<len\_12>}; 
(2) parts of the lyric text are randomly replaced with \texttt{<mask>} tokens. 
The length tokens and masked lyrics are concatenated into one sequence and passed to 
a Transformer encoder–decoder, which attempts to reconstruct the original unmasked 
lyrics. Arrows illustrate the flow from original text to masking, length extraction, 
concatenation, and reconstruction.
}
% \vskip -5mm
\label{fig:unsupervised}
\end{center}
\end{figure}

% Training procedure
This self-supervised training stage, as illustrated in Figure \ref{fig:unsupervised}, serves two purposes simultaneously. First, we continue BART’s original mask-infilling objective \citep{lewis2020bart}: randomly selected spans of the input text are masked, and the model is trained to reconstruct the missing content. This allows the model to further adapt its generation behavior toward the stylistic characteristics of lyrics.

Second, we prepend the length-control tokens described in the previous subsection to the masked input. These tokens specify, for each line, the target number of syllables the model should reproduce. By conditioning the reconstruction task on these prompts, the model learns to associate the length tokens with concrete syllable counts, thereby developing strong length awareness before paired melody--lyric training begins.

This design allows the subsequent supervised M2L fine-tuning stage to focus on more fine-grained melody–lyric compatibility rather than struggling with basic length alignment.

\subsection{Enhancing Finer-Grained Compatibility}

Beyond length alignment, melody--lyric compatibility depends on subtler prosodic relationships. As discussed in Section~\ref{sec:related_singability}, prior analyses---notably \citet{nichols2009relationships}---show that properties such as word importance, syllable stress, and vowel duration correlate with musical attributes like note duration, metric position, and melodic peaks. These relationships are flexible rather than deterministic, yet they collectively shape perceived singability. Our goal is therefore to introduce training mechanisms that encourage the model to internalize these correlations and to reflect them during generation.

\subsubsection{Joint Learning of Output Formatting}
\label{sec:joint}

% Motivation
Although self-supervised training establishes length awareness, supervised melody-to-lyric fine-tuning is still helpful for learning finer compatibility patterns. Because the rules linking melody and lyrical prosody are complex, variable, and difficult to express explicitly, we rely on the model to learn them implicitly. However, as demonstrated in Section~\ref{sec:ablation}, the limited paired data available for M2L restrict the model’s ability to infer such patterns solely from standard sequence-to-sequence training. 

% Method
To strengthen the encoder’s ability to capture prosodically meaningful cues, we introduce auxiliary supervision that explicitly links melody positions to prosodic attributes of their aligned syllables. For each auxiliary task, we attach a position-wise linear classifier on top of the melody encoder to predict whether a particular lyric attribute will happen at the syllable correspond to each note position. The intuition is that the auxiliary tasks encourage $h_j$ to encode structural and prosodic information that is essential for singability, thus providing the decoder with richer and more reliable conditioning.

\begin{table}[tb]
\tbl{Labels for classification tasks.}
{
\begin{tabular}{@{}lll@{}}
\toprule
\textbf{Classification Task}           & \textbf{Label} & \textbf{Meaning}                              \\ \midrule
\multirow{3}{*}{Syllable stress}       & 0              & Unstressed syllables                          \\
                                       & 1              & Syllables with primary stress                 \\
                                       & 2              & Syllables with secondary stress               \\ \midrule
\multirow{3}{*}{Word importance level} & 0              & Stop words                                    \\
                                       & 1              & Non-stop words with lower 50\% TF-IDF scores  \\
                                       & 2              & Non-stop words with higher 50\% TF-IDF scores \\ \midrule
\multirow{3}{*}{Vowel duration}            & 0              & Short vowels                                  \\
                                       & 1              & Long vowels                                   \\
                                       & 2              & Diphthongs                                    \\ \bottomrule
\end{tabular}
}
% \caption{Labels for classification tasks.}
\label{tab:cls_labels}
\end{table}

% \begin{table}
% \tbl{Example of a table showing that its caption is as wide as
% the table itself and justified.}
% {\begin{tabular}{lcccccc} \toprule
% & \multicolumn{2}{l}{Type} \\ \cmidrule{2-7}
% Class & One & Two & Three & Four & Five & Six \\ \midrule
% Alpha\textsuperscript{a} & A1 & A2 & A3 & A4 & A5 & A6 \\
% Beta & B2 & B2 & B3 & B4 & B5 & B6 \\
% Gamma & C2 & C2 & C3 & C4 & C5 & C6 \\ \bottomrule
% \end{tabular}}
% \tabnote{\textsuperscript{a}This footnote shows how to include
% footnotes to a table if required.}
% \label{sample-table}
% \end{table}

% \usepackage{booktabs}
% \usepackage{graphicx}
\begin{table}[]
\centering
\tbl{Example of label extraction for the syllable attribute classification tasks. Original sentence: "The snow glows white on the mountain tonight".}{

\resizebox{\columnwidth}{!}{%
\begin{tabular}{@{}cllrccc@{}}
\toprule
\multicolumn{1}{l}{} &  & \multicolumn{2}{c}{\textbf{Word Attribute}} & \multicolumn{3}{c}{\textbf{Syllable Attributes}} \\
\multicolumn{1}{l}{\textbf{Syllable ID}} & \textbf{Word} & \textbf{ARPABET Phonemes} & \textbf{TF-IDF} & \multicolumn{1}{l}{\textbf{Stress}} & \multicolumn{1}{l}{\textbf{Importance}} & \multicolumn{1}{l}{\textbf{Vowel duration}} \\ \midrule
1 & The & {[}DH, AH0{]} & (stop word) & 0 & 0 & 0 \\
2 & snow & {[}S, N, OW1{]} & 6.21 & 1 & 2 & 2 \\
3 & glows & {[}G, L, OW1, Z{]} & 7.60 & 1 & 2 & 2 \\
4 & white & {[}W, AY1, T{]} & 3.91 & 1 & 1 & 2 \\
5 & on & {[}AA1, N{]} & (stop word) & 1 & 0 & 1 \\
6 & the & {[}DH, AH0{]} & (stop word) & 0 & 0 & 0 \\
7 & mountain & {[}M, AW1, N, & 5.30 & 1 & 2 & 2 \\
8 &  & T, AH0, N{]} & - & 0 & 2 & 0 \\
9 & tonight & {[}T, AH0, & 4.61 & 0 & 1 & 0 \\
10 &  & N, AY1, T{]} & - & 1 & 1 & 2 \\ \bottomrule
\end{tabular}%
}
}
\label{tab:label_eg}
\end{table}

For each melody note position $j$, we define three auxiliary prediction tasks corresponding to the prosodic attributes of the lyric syllable aligned to that note:
\begin{itemize}
    \item \textbf{Syllable stress prediction:} predict the stress category (primary, secondary, or unstressed) of the syllable aligned to note $j$.
    \item \textbf{Word-importance prediction:} determine whether the aligned syllable belongs to a low-importance word (e.g., stopwords), a medium-importance word, or a high-importance word.
    \item \textbf{Vowel-duration prediction:} predict the vowel length category (short, medium, long) of the syllable aligned to note $j$.
\end{itemize}
These tasks are performed at the encoder side, with one classifier output per melody note, enabling the encoder to learn representations that capture prosodic cues relevant for singable lyric generation.

\paragraph*{Label Construction.}
Each classification problem is formulated as a three-class prediction. Table~\ref{tab:cls_labels} summarizes the definitions of the label categories. Although the original paired dataset contains no such labels, they can be derived automatically from the ground-truth lyrics. Syllable stress and vowel categories are obtained from CMUDict, combined with a simplified vowel–duration mapping (Appendix~\ref{app:vowel_len}) to get vowel duration. Word importance is determined using stopword identification and TF--IDF scores \citep{sparck1972statistical}. An example of this label derivation process is shown in Table~\ref{tab:label_eg}.

\paragraph*{Training Objective.}
The three classifiers are trained jointly with the sequence-to-sequence M2L fine-tuning, using cross-entropy objectives. The total loss is:
\begin{equation}
\label{eq:loss}
L = \text{CE}(\mathbf{y}, \hat{\mathbf{y}}) 
\;+\;
\frac{1}{|\mathbf{y}|}
\sum_{j=1}^{|\mathbf{y}|}
\Big[
    \text{CE}(s_j, \hat{s}_j)
    +
    \text{CE}(i_j, \hat{i}_j)
    +
    \text{CE}(v_j, \hat{v}_j)
\Big],
\end{equation}
where $\mathbf{y}$ is the target lyric sequence; $s_j$, $i_j$, and $v_j$ denote the stress, importance, and vowel-duration labels at position $j$; and hats denote predictions. Length prompts remain part of the input during this stage, preserving the length-awareness acquired from self-supervised training to encourage the model to focus on more fine-grained compatibility cues.

\subsubsection{Handling Music Input}

As defined in our problem formulation, each input note is described by three attributes: onset, offset, and pitch, where onset and offset are expressed in seconds. However, directly using absolute timing in seconds often yields a \emph{sparse} distribution of temporal values—that is, note onsets and durations occur at irregular, widely separated numeric scales, resulting in large gaps and very few repeated values. Such sparsity can hinder the model from learning temporal patterns in melodies.

\begin{figure}[tb]
\begin{center}
\includegraphics[width=0.6\columnwidth]{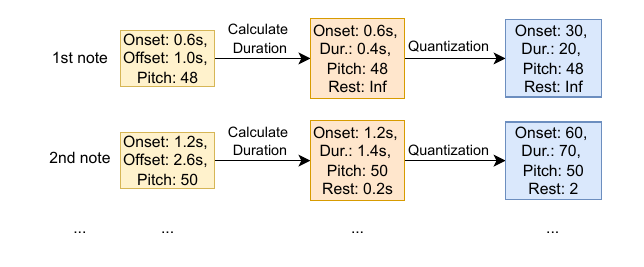}
\caption{\textbf{Caption:} Illustration of the melody preprocessing procedure. Each note is first represented by its onset, offset, and pitch. Durations and inter-note rests are then computed, followed by quantization into discretized onset, duration, pitch, and rest tokens used as model inputs. \\\hspace{\textwidth} \textbf{Figure 4. Alt Text}: The figure illustrates the preprocessing steps applied to melody notes before model input. For the first note, the raw onset, offset, and pitch (0.6s, 1.0s, 48) are used to compute a duration of 0.4s and an infinite rest value. These values are then quantized into discrete tokens (onset 30, duration 20, pitch 48, rest Inf). For the second note, (1.2s, 2.6s, 50) becomes a duration of 1.4s and a rest of 0.2s, which are quantized into (onset 60, duration 70, pitch 50, rest 2). The diagram shows this process repeating for subsequent notes.}
\label{fig:mel_eg}
\end{center}
\end{figure}

% \caption{Illustration of the melody preprocessing procedure. Each note is first represented by its onset, offset, and pitch. Durations and inter-note rests are then computed, followed by quantization into discretized onset, duration, pitch, and rest tokens used as model inputs.}

% \alttext{The figure illustrates the preprocessing steps applied to melody notes before model input. 
% For the first note, the raw onset, offset, and pitch (0.6s, 1.0s, 48) are used to compute a duration of 0.4s and an infinite rest value. These values are then quantized into discrete tokens (onset 30, duration 20, pitch 48, rest Inf). 
% For the second note, (1.2s, 2.6s, 50) becomes a duration of 1.4s and a rest of 0.2s, which are quantized into (onset 60, duration 70, pitch 50, rest 2). The diagram shows this process repeating for subsequent notes.}

% Quantization
Hence we normalize and quantize time-related note attributes to alleviate such sparsity issue and remove tempo-induced variation across songs. The melody is ultimately converted into a sequence of note-level embeddings that the encoder can process. Pitch values are quantized to their nearest semi-tone, represented by integer 1--128. For the temporal attributes, we apply the following procedure.

Within each paragraph, we take the duration of the shortest note as the reference duration unit, denoted $\text{dur}_u$. All onset and duration values are then divided by $\text{dur}_u$, multiplied by~10, and floored to achieve a resolution of 0.1 times reference duration unit. In addition, each onset is adjusted by subtracting the first onset in the paragraph. Beyond note duration, we introduce an additional variable, $\text{rest}_j$, representing the length of the silence preceding the $j$-th note.

% This yields a more compact and learnable temporal representation. 

The full quantization process is shown in Equations~\eqref{eq:dur_u}--\eqref{eq:rest_q}:
\begin{align}
    \text{dur}^{(s)}_u &= \min_{1 \le j \le |y|}\left\{\text{off}^{(s)}_j - \text{on}^{(s)}_j\right\},
        \label{eq:dur_u} \\[6pt]
    \text{on}^{(q)}_j &= 
        \min\!\left(
        \left\lfloor 10 \cdot \frac{\text{on}^{(s)}_j - \text{on}^{(s)}_1}{\text{dur}^{(s)}_u} \right\rfloor,
        \text{vocab}_{\text{on}}
        \right),
        \quad \forall j,
        \label{eq:on_q} \\[6pt]
    \text{dur}^{(q)}_j &= 
        \min\!\left(
        \left\lfloor 
            10 \cdot \frac{\text{off}^{(s)}_j - \text{on}^{(s)}_j}{\text{dur}^{(s)}_u}
        \right\rfloor,
        \text{vocab}_{\text{dur}}
        \right),
        \quad \forall j,
        \label{eq:dur_q} \\[6pt]
    \text{rest}^{(q)}_j &= 
    \begin{cases}
        \inf, & j = 1, \\
        \min\!\left(
            \left\lfloor 
                10 \cdot \frac{\text{on}^{(s)}_j - \text{off}^{(s)}_{j-1}}{\text{dur}^{(s)}_u}
            \right\rfloor,
            \text{vocab}_{\text{rest}}
        \right), & j \ge 2,
    \end{cases}
    \label{eq:rest_q}
\end{align}
where variables with superscript $(s)$ denote values in seconds, and those with $(q)$ denote the quantized model inputs. An illustration of this processing pipeline is shown in Figure~\ref{fig:mel_eg}.

% Compound embedding
To represent each quantized melody paragraph in the model, we allocate three discrete token vocabularies corresponding to onsets, durations, and rest intervals. Their sizes are bounded by $\text{vocab}_{\text{on}}$, $\text{vocab}_{\text{dur}}$, and $\text{vocab}_{\text{rest}}$, set to 640, 640, and 240 in our experiments (covering 64, 64, and 24 times of reference duration unit, respectively). Each token type is embedded using a separate embedding layer, and the resulting vectors are summed to form a compound embedding for each note. All embedding dimensions match the hidden size of the BART encoder to ensure compatibility.

% \subsection{System Overview}

\begin{figure}[tb]
\begin{center}
\includegraphics[width=1.0\linewidth]{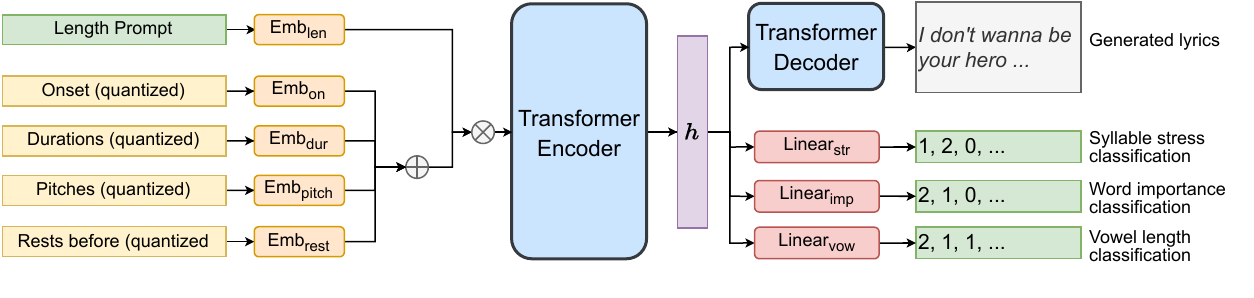}
\caption{\textbf{Caption}: Overview of the supervised melody--to--lyric (M2L) training stage. Quantized note attributes (onset, duration, pitch, and rest) are embedded and summed to form note representations, which are concatenated with length-prompt embeddings before entering the Transformer encoder. The encoder output $h$ is fed into the Transformer decoder for lyric generation, while three auxiliary linear heads predict syllable stress, word importance, and vowel length for each note position. \\\hspace{\textwidth} \textbf{Figure 5. Alt Text}: A block diagram illustrating the supervised melody-to-lyric training architecture. 
On the left, four types of quantized note attributes—onsets, durations, pitches, and rest intervals—each pass through their own embedding layer. These embeddings are summed to form a composite note embedding sequence. A separate embedding stream encodes the length prompts. The length-prompt embeddings and note embeddings are concatenated and fed into a Transformer encoder. The encoder outputs a hidden sequence $h$, which is passed to a Transformer decoder to generate the lyric text. In parallel, three linear classification heads attached to the encoder predict, for each note position, the syllable stress class, word importance class, and vowel length class. The diagram visually emphasizes the joint learning of lyric generation and formatting predictions.}
% \vskip -5mm
\label{fig:system}
\end{center}
\end{figure}

The overall architecture of our model is shown in Fig.~\ref{fig:system}. The system follows a Transformer-based encoder--decoder framework, with inputs drawn from two sources: (1) \emph{length prompts}, which specify the required syllable count for each lyric line, and (2) \emph{melody note attributes}, including quantized onset, duration, pitch, and rest intervals. Each note attribute is mapped through its own embedding layer, and the resulting embeddings are summed to obtain a single note representation. The sequence of length prompts is then concatenated with the sequence of note embeddings to form the final encoder input.

The decoder generates the lyric paragraph conditioned on these enriched encoder representations. Combined with the learning strategies introduced in the previous subsections, this architecture enables the model to produce well-formed, singable lyrics aligned with the melody. An example output is shown in the fourth line of Fig.~\ref{fig:intro}, and additional case studies are presented in \S\ref{sec:case}.

\section{Experiments}

This section describes the experimental setup used to evaluate the proposed melody-to-lyric generation framework. 

\subsection{Model Details}

% Model structure 
We adopt BART-base\footnote{\url{https://huggingface.co/facebook/bart-base}} \citep{lewis2020bart} as the backbone of our system. This model, pretrained on large-scale general-domain text corpora, provides a compact yet capable foundation for evaluating our training strategy. Since the focus of this work is to examine the effectiveness of length-aware training and formatting supervision—rather than to maximize absolute generation quality via scaling—a lightweight architecture such as BART-base allows clearer attribution of performance gains and reduces computational cost during experimentation.

% Hparam
Our training proceeds in two stages: (1) self-supervised length-aware learning on text-only lyric data, and (2) supervised fine-tuning on paired melody–lyric data. For both stages, the batch size is set to the maximum compatible with a single NVIDIA A5000 GPU (24 GB), which is 48 for both text-only and paired-data settings. A simple learning rate search yields $2 \times 10^{-4}$ for the self-supervised stage and $1 \times 10^{-4}$ for the supervised stage. We employ the AdamW optimizer \citep{DBLP:conf/iclr/LoshchilovH19}, with a linear learning-rate decay schedule for the self-supervised phase. Warm-up steps are set to 2{,}500 for self-supervised training and 200 for supervised training. The model is trained for 15 epochs on the text-only corpus and 10 epochs on the paired corpus. For evaluation, we select the checkpoint with the lowest validation loss.

\subsection{Baseline Systems}

We compare our model's performance with several external baselines, include two task-specific baseline: SongMASS \citep{sheng2021songmass} and AI-Lyricist \citep{ma2021ai}. Additionally, ChatGPT~5.1 \citep{openai_chatgpt51} is included as an LLM-based baseline in objective evaluation.

For ablation studies, we compare the full model against several variants in which one or more proposed components are removed, while keeping other training conditions identical. This design isolates the contribution of each component and allows a fair assessment of how the individual elements of our training strategy and formatting supervision influence overall performance. We will explain model configurations in Section \ref{sec:ablation}.

\subsection{Dataset}

\begin{table}[tb]
\tbl{Dataset size of different splits.}
{
\begin{tabular}{@{}ll|rrrr@{}}
\toprule
                                                     &              & \textbf{Train} & \textbf{Validation} & \textbf{Test} & \textbf{Total} \\ \midrule
\multirow{2}{*}{\textbf{Text-only}}                  & \#paragraphs & 519,616        & 1,993               & 1,996         & 523,605        \\
                                                     & \#lines      & 7,046,894      & 26,684              & 27,109        & 7,100,687      \\
\multicolumn{1}{c}{\multirow{2}{*}{\textbf{Paired}}} & \#paragraphs & 47,222         & 1,989               & 1,989         & 51,200         \\
\multicolumn{1}{c}{}                                 & \#lines      & 279,682        & 11,667              & 11,430        & 302,779        \\ \bottomrule
\end{tabular}%
}
\label{tab:data}
\end{table}

% Raw data
We use DALI v2 \citep{meseguer2020creating} as the paired melody--lyric dataset due to its relatively high annotation quality and alignment accuracy. The dataset provides not only melody annotations but also line and paragraph boundaries, which are essential for paragraph-level M2L generation. For the text-only lyric corpus, we adopt a large multilingual lyric collection from Kaggle,\footnote{\url{https://www.kaggle.com/datasets/mateibejan/multilingual-lyrics-for-genre-classification}} which is substantially larger than the paired set and thus well-suited for length-aware self-supervised training.

We note that many previous works rely on the dataset from \cite{yu2021conditional}, originally designed for the lyric-to-melody (L2M) task. However, our inspection revealed numerous alignment issues—misaligned lyrics, missing words, and inconsistent boundaries—making it suboptimal for generating high-quality lyrics aligned to melody. For this reason, we did not adopt it as the training corpus.

% Preprocessing
Before training, we apply text normalization to both datasets: removing non-English items, converting to lowercase, stripping special symbols, and removing blank lines. We also perform paragraph-level deduplication, removing repeated sections such as duplicated choruses. This is motivated by findings showing that deduplication improves training efficiency, generation diversity, and evaluation reliability of language models \citep{lee2022deduplicating}. We then create train/validation/test by random spliting and ensure no paragraph overlap across splits. Dataset statistics for all splits are listed in Table~\ref{tab:data}.\footnote{We will release the processed dataset upon acceptance to support reproducibility.}

The CMU Pronunciation Dictionary\footnote{\url{http://www.speech.cs.cmu.edu/cgi-bin/cmudict}} and NLTK \citep{loper2002nltk} are used to derive the classification labels described in Section~\ref{sec:joint}, including syllable stress, vowel type, and stopword identification.

% Separate evaluation dataset
Finally, we use a separate out-of-domain test set for a fair comparison with external baseline systems, since these baselines are adopted as-is without retraining. We select 11 paragraphs from six randomly chosen songs in the \cite{yu2021conditional} dataset. This provides an out-of-domain evaluation setting for both objective metrics and human evaluation, while the in-domain test set is reserved for ablation studies.

\subsection{Objective Metrics}

We compare our model against several external baselines using both objective metrics and human evaluation, with the original human-created lyrics serving as the upper-performance reference in all assessments.

For objective evaluation, we examine text quality together with both coarse- and fine-grained measures of melody–lyric compatibility. 

\subsubsection{Text Quality}

Because our system does not impose content constraints, the generated lyrics for a given melody that satisfy the implicit format requirement may differ substantially in contents and wording from the ground truth. Therefore, similarity-based metrics such as BLEU \citep{papineni2002bleu} or ROUGE \citep{lin2004rouge}---which assume that outputs closer to the reference indicate higher quality---are not suitable for this task.  

Instead, we assess textual quality using perplexity, computed by evaluating each model on the test set. Lower perplexity indicates that the generated text is more consistent with human-written language and thus reflects higher linguistic quality (and, indirectly, better compatibility). To ensure a fair comparison, perplexity is only applied in the ablation study, where all systems share the same architecture and training data.

\subsubsection{Syllable Alignment}
We evaluate whether the generated lyrics satisfy the basic structural constraints of melody–lyric alignment. Two metrics are used:

\begin{enumerate}
    \item \textbf{\#Line}: The paragraph-level accuracy of line count, i.e., the proportion of generated paragraphs whose number of lines matches the required number of melody phrases.
    \item \textbf{Line len}: The average accuracy of syllable count per line, evaluating how reliably the model produces the correct number of syllables for each melodic phrase.
\end{enumerate}

These two metrics measure whether the generated paragraphs meet the fundamental one-to-one note--syllable alignment constraints defined in the task.

\subsubsection{Fine-Grained Compatibility}

To assess music--lyric compatibility beyond length, we employ sentence-level metrics that quantify how often specific melodic properties co-occur with relevant syllabic properties. These metrics are based on the \emph{coexistence probability}.

Let $\alpha$ denote a property of a musical note and $\beta$ a property of the corresponding lyric syllable. The coexistence probability between $\alpha$ and $\beta$ is defined as:

\begin{align}
\Pr(\alpha\text{-}\beta) &= \frac{1}{|X|}\sum_{\{(x,y) \mid |x| = |y|\}}
\Pr(\alpha\text{-}\beta; x, y),
\label{eq:joint} \\
\Pr(\alpha\text{-}\beta; x, y) &=
\frac{\text{JointCount}(\alpha, \beta; x, y)}
     {\text{Count}(\alpha; x)},
\end{align}

where:
\begin{itemize}
    \item $(x,y)$ is a melody–lyric pair of equal length,
    \item $\text{Count}(\alpha; x)$ is the number of notes in $x$ exhibiting property $\alpha$,
    \item $\text{JointCount}(\alpha, \beta; x, y)$ counts aligned note–syllable pairs $(x_i,y_i)$ where $x_i$ has $\alpha$ and $y_i$ has $\beta$,
    \item $|X|$ denotes the number of evaluated examples.
\end{itemize}

A high value of $\Pr(\alpha\text{-}\beta)$ (close to 1) indicates that the note property $\alpha$ and syllable property $\beta$ consistently appear at corresponding positions—i.e., they co-occur wherever $\alpha$ appears. Human-written lyrics exhibit characteristic coexistence patterns across several $\alpha\text{-}\beta$ pairs. We compute these values on the ground truth and treat them as desirable targets. A model that produces coexistence probabilities closer to human data is assumed to exhibit higher music–lyric compatibility and better singability. In contrast, randomly generated text rarely exhibits strong co-occurrence.

We apply Eq.~\ref{eq:joint} to the following five note–syllable property pairs:

\begin{enumerate}
    \item \textbf{Dur-str}: long notes (top 50\% duration) vs.\ stressed syllables (primary or secondary stress)\footnote{Although the duration threshold is a simplification, it is applied consistently across all systems and evaluations.}
    \item \textbf{Peak-str}: melodic peaks (notes with pitches higher than both neighbors) vs.\ stressed syllables
    \item \textbf{Dur-imp}: long notes vs.\ syllables belonging to important words (non-stopwords)
    \item \textbf{Peak-imp}: melodic peaks vs.\ important words
    \item \textbf{Dur-vow}: long notes vs.\ syllables with long vowels (long vowels or diphthongs)
\end{enumerate}

As highlighted by \cite{nichols2009relationships}, these properties reflect common correlations in human-composed songs.

\subsection{Human Evaluation}

While objective metrics provide useful diagnostic signals, they cannot fully capture the qualitative dimensions of lyric generation. Improvements in fine-grained compatibility, for example, may occasionally come at the cost of reduced textual fluidity. Since no single automatic metric—or combination of metrics—can reliably assess this trade-off, human evaluation is indispensable.

We recruited ten university students with experience in amateur-level music performance or lyric writing. Each participant evaluated multiple lyric versions for each paragraph, assigning a score on a 5-point Likert scale (5 = very good, 1 = very bad). The final score for each lyric paragraph was computed as the average across participants. Evaluations were conducted along three dimensions:

\begin{enumerate}
    \item \textbf{Fluency}: Grammaticality and semantic coherence of the lyric text.
    \item \textbf{Music--Lyric Compatibility}: The degree to which the lyrics align with the melody and the ease with which they can be sung.
    \item \textbf{Overall Quality}: A holistic judgment reflecting both linguistic quality and musical compatibility, representing the central goal of singable lyric generation.
\end{enumerate}

All evaluations were carried out at the paragraph level. For each paragraph, six versions (one original and five generated) were shown simultaneously to facilitate direct comparison. Evaluators were blinded to the identities and origins of all systems.

To provide sufficient musical context, participants were given the corresponding music score alongside synthesized singing audio. The audio consisted of a vocal track synthesized to match the original melody and each lyrical variant, mixed with the original accompaniment. The singing voice was synthesized using ACE Studio,\footnote{\url{https://ace-studio.huoyaojing.com/}} while accompaniment tracks were extracted using the Demucs v3 \texttt{mdx\_extra} source separation model \citep{defossez2019demucs}.
\section{Results}

% Section intro
This section presents the empirical results of our study. We first compare model aginst other baselins on both objective metrics and human evaluation. Then we show results of ablation study to show contribution of individual model components. Finally, we provide a qualitative case study to illustrate the strengths and remaining limitations of our system through representative examples.

\subsection{Objective Evaluation}
\label{sec:res_obj}

Table~\ref{tab:objective} summarizes the objective comparison between our system and the external baselines. The key observations are as follows.

% Lenght control sucks
\textbf{Baseline systems struggle to satisfy basic length constraints.}  
Both SongMASS and AI-Lyricist show clear limitations in controlling the number of lyric lines and the syllable count per line. Their generated outputs frequently violate these basic structural requirements, indicating inconsistent enforcement of length constraints. ChatGPT performs no better; in fact, it shows particularly poor control over syllable counts (only 6.67\% accuracy), suggesting that it lacks the ability to reliably integrate phonological and formatting constraints. In contrast, our model achieves perfect accuracy on the number of lines (100\%) and near-perfect control of syllable counts (98.33\%). This strong length adherence provides a stable foundation for modeling more advanced aspects of formatting and compatibility.

% Ours is significantly better at fine-grained compatibility
\textbf{Our model consistently outperforms baselines in fine-grained compatibility metrics.}  
Across all evaluated melody--lyric compatibility metrics, our system exhibits patterns that closely resemble those found in human-written lyrics. For example, in the reference lyrics, long-duration notes co-occur with stressed syllables 73.44\% of the time. Our model achieves a comparable 70.46\%, whereas SongMASS and AI-Lyricist remain below 20\%. Similar trends hold for all other property pairs. ChatGPT again does not outperform the other baselines, reflecting its limited capability of handling prosodic alignment. These results show that our system not only satisfies length constraints but also captures more subtle prosodic correspondences that the baselines fail to learn.

% Coocurrance is dynamic
\textbf{Melody and lyric properties are not deterministically coupled.}  
The the absolute values of compatibility metrics calcluated over human-created lyrics differ across evaluation datasets (e.g., between the out-of-domain evaluation here and the in-domain ablation study in Table~\ref{tab:ablation}). This variation reflects the fact that melody--lyric relationships are flexible and song-dependent rather than rigid rules. Despite this variation, our model consistently exhibits the highest similarity to human co-occurrence patterns across all experimental settings.

% \usepackage{booktabs}
% \usepackage{graphicx}
\begin{table}[tb]
\centering

\tbl{The main result of objective evaluation. The best results are \textbf{bolded}. We report mean and standard deviation of mean (SEM).}
{
% \resizebox{\columnwidth}{!}{%
\begin{tabular}{lrrrrrrr}
\hline
\textbf{Model} & \textbf{\#Line} & \textbf{Line Len} & \textbf{Dur-str} & \textbf{Peak-str} & \textbf{Dur-imp} & \textbf{Peak-imp} & \textbf{Dur-vow} \\ \hline
Original lyrics & - & - & 73.44 & 50.27 & 25.46 & 19.81 & 23.38 \\ \hline
Ours & \textbf{100.00} & \textbf{98.33} & \textbf{70.46} & \textbf{68.40} & \textbf{19.32} & \textbf{16.34} & \textbf{28.01} \\
SongMASS & 81.82 & 15.00 & 14.14 & 18.18 & 3.48 & 7.58 & 4.49 \\
AI-Lyricist & 72.73 & 28.33 & 19.08 & 11.09 & 4.85 & 4.11 & 4.83 \\
ChatGPT & 72.73 & 6.67 & 5.41 & 9.63 & 4.15 & 8.12 & 1.30 \\ \hline
\end{tabular}%
}
% }
\label{tab:objective}
\end{table}

\subsection{Human Evaluation}
\label{sec:subjective}

\begin{table}[tb]
\tbl{Subjective evaluation results. The best results are \textbf{bolded}.}
{
\begin{tabular}{@{}llll@{}}
\toprule
 & Fluency & Compatibility & Overall Quality \\ \midrule
Original & 4.00 ± 0.13 & 4.24 ± 0.12 & 4.15 ± 0.12 \\ \midrule
AI-Lyricist & 2.20 ± 0.09 & 2.40 ± 0.10 & 2.21 ± 0.09 \\
SongMASS & 1.70 ± 0.09 & 1.95 ± 0.09 & 1.82 ± 0.10 \\ \midrule
BART ft. & 2.97 ± 0.12 & 2.24 ± 0.11 & 2.39 ± 0.09 \\
Self-supervised & \textbf{3.42} ± 0.12 & 3.06 ± 0.10 & 3.07 ± 0.10 \\
Ours & \textbf{3.42} ± 0.12 & \textbf{3.15} ± 0.09 & \textbf{3.15} ± 0.09 \\ \bottomrule
\end{tabular}%
}
\label{tab:human_eval}
\end{table}

% Baseline is bad
Table~\ref{tab:human_eval} summarizes the results of the subjective evaluation. Compared with the two external task-specific baselines, our final model achieves the highest scores across all three criteria---fluency, melody--lyric compatibility, and overall quality---and does so by a substantial margin. SongMASS \citep{sheng2021songmass}, despite being trained on the same corpus from which the test set is drawn, receives the lowest ratings among all systems. AI-Lyricist \citep{ma2021ai} performs noticeably better than SongMASS but remains clearly inferior to both our baseline and full models.

% 2-stage training is useful
Ablation results further highlight the benefit of the proposed two-stage training strategy. Using only supervised sequence-to-sequence fine-tuning (BART ft.) does not yield a clear advantage over the AI-Lyricist baseline in either compatibility or overall quality. In contrast, applying only the self-supervised pretraining stage provides improvements across all metrics, likely due to the expanded effective training set. Performing supervised fine-tuning after the self-supervised stage produces additional gains, particularly for melody--lyric compatibility and overall quality, showing the effectiveness of the proposed 2-stage training.

\subsection{Ablation Study}
\label{sec:ablation}

% \usepackage{booktabs}
\begin{table}[tb]
\tbl{Model configurations in the ablation study.}
{
\resizebox{\columnwidth}{!}{

\begin{tabular}{@{}llllcc@{}}
\toprule
\textbf{Model Name} & \multicolumn{1}{c}{\textbf{BART ft.}} & \multicolumn{1}{c}{\textbf{Adapted}} & \multicolumn{1}{c}{\textbf{Self-supervised}} & \textbf{Single-task} & \textbf{Ours} \\ \midrule
In-domain pretraining &  & \multicolumn{1}{c}{\checkmark} & \multicolumn{1}{c}{\checkmark} & \checkmark & \checkmark \\
Length prompt in self-supervised training &  &  & \multicolumn{1}{c}{\checkmark} & \checkmark & \checkmark \\
M2L fine-tuning & \multicolumn{1}{c}{\checkmark} & \multicolumn{1}{c}{\checkmark} &  & \checkmark & \checkmark \\
Length prompt in M2L fine-tuning &  &  &  & \checkmark & \checkmark \\
Auxiliary classification supervision &  &  &  & \multicolumn{1}{l}{} & \checkmark \\ \bottomrule
\end{tabular}%

}
}
\label{tab:models}
\end{table}
% \usepackage{booktabs}
\begin{table}[tb]
\tbl{Results of the ablation study with objective evaluation. The best results are in \textbf{bold}.}
{
\resizebox{\columnwidth}{!}{

\begin{tabular}{@{}clcccccccc@{}}
\toprule
\textbf{No.} & \multicolumn{1}{c}{\textbf{Model}} & \textbf{PPL} & \textbf{\#Line} & \textbf{Line len} & \textbf{Dur-str} & \textbf{Peak-str} & \textbf{Dur-imp} & \textbf{Peak-imp} & \textbf{Dur-vow} \\ \midrule
- & Original lyrics & - & 100.00 & 100.00 & 82.45 & 66.45 & 72.29 & 52.73 & 59.99 \\ \midrule
1 & BART ft. & 10.84 & 95.40 & 75.67 & 64.61 & 47.96 & 45.94 & 30.30 & 50.10 \\
2 & Adapted & 10.98 & 14.28 & 13.15 & 10.44 & 8.88 & 8.26 & 5.08 & 7.20 \\
3 & Self-supervised & 7.99 & 99.35 & 97.15 & 75.25 & 58.99 & \textbf{54.78} & 41.18 & 54.06 \\
4 & Single-task & 7.96 & \textbf{99.65} & \textbf{97.42} & 75.15 & 59.91 & 54.44 & \textbf{41.49} & 53.75 \\
5 & Ours & \textbf{7.93} & 99.15 & 97.11 & \textbf{77.78} & \textbf{61.36} & 54.33 & 40.04 & \textbf{55.96} \\ \bottomrule
\end{tabular}%

% \begin{tabular}{@{}clcccccccc@{}}
% \toprule
% \textbf{No.} & \multicolumn{1}{c}{\textbf{Model}} & \textbf{PPL} & \textbf{\#Line} & \textbf{Line len} & \textbf{Dur-str} & \textbf{Peak-str} & \textbf{Dur-imp} & \textbf{Peak-imp} & \textbf{Dur-vow} \\ \midrule
% - & Original lyrics & - & 100.00 & 100.00 & 82.45 & 66.45 & 72.29 & 52.73 & 59.99 \\ \midrule
% 1 & BART ft. & 10.84 & 95.40 & 75.67 & 64.61 & 47.96 & 45.94 & 30.30 & 50.10 \\
% 2 & Adapted & 10.98 & 14.28 & 13.15 & 10.44 & 8.88 & 8.26 & 5.08 & 7.20 \\
% 3 & Unsupervised & 7.99 & 99.35 & 97.15 & 75.25 & 58.99 & \textbf{54.78} & 41.18 & 54.06 \\
% 4 & Supervised & 11.62 & 95.10 & 70.67 & 57.86 & 42.33 & 38.97 & 26.74 & 43.28 \\
% 5 & Single-task & 7.96 & \textbf{99.65} & \textbf{97.42} & 75.15 & 59.91 & 54.44 & \textbf{41.49} & 53.75 \\
% 6 & Ours & \textbf{7.93} & 99.15 & 97.11 & \textbf{77.78} & \textbf{61.36} & 54.33 & 40.04 & \textbf{55.96} \\ \bottomrule
% \end{tabular}%
}
}
\label{tab:ablation}
\end{table}

Table~\ref{tab:models} summarizes the configurations of our ablation variants, and Table~\ref{tab:ablation} reports their objective performance. Here are some observations.

\textbf{Our approach on format control does not harm text quality.}
Across all variants, our full model (No.~4) achieves the lowest perplexity, indicating the highest text quality. This enhancement is likely attributed to these constraints acting as supplementary guidance during training, enabling the model to learn the melody-specific lyric style more effectively.

\textbf{Additional supervised training offers limited gains unless the auxiliary supervision is included.}
Under the two-stage setting, adding supervised M2L fine-tuning without the auxiliary classifiers (No.~4) yields performance that is comparable to the self-supervised-only model (No.~3). When the auxiliary supervision is added (No.~5), the model shows modest but consistent improvements on several fine-grained compatibility metrics (dur–str +2.63\%, peak–str +1.45\%, dur–vow +2.21\%), one metric remains essentially unchanged (dur–imp –0.11\%), and one decreases slightly (peak–imp –1.45\%). These results indicate that the auxiliary objectives do not dramatically alter performance, but they provide incremental benefits that help the model capture certain prosodic tendencies more reliably.

\textbf{Early incorporation of length prompts is crucial for learning length awareness.}
If length prompts are not introduced during the self-supervised stage (No.~2), the subsequent supervised M2L training fails to achieve length control. In contrast, when length prompts are included from the beginning (No.~3), the model acquires strong length awareness even without supervised training. This confirms that length prompts must be embedded early in training to ensure reliable control of both line count and syllable count.

\subsection{Case Study}
\label{sec:case}

\begin{figure}[tb]
\begin{center}
\includegraphics[width=0.9\columnwidth]{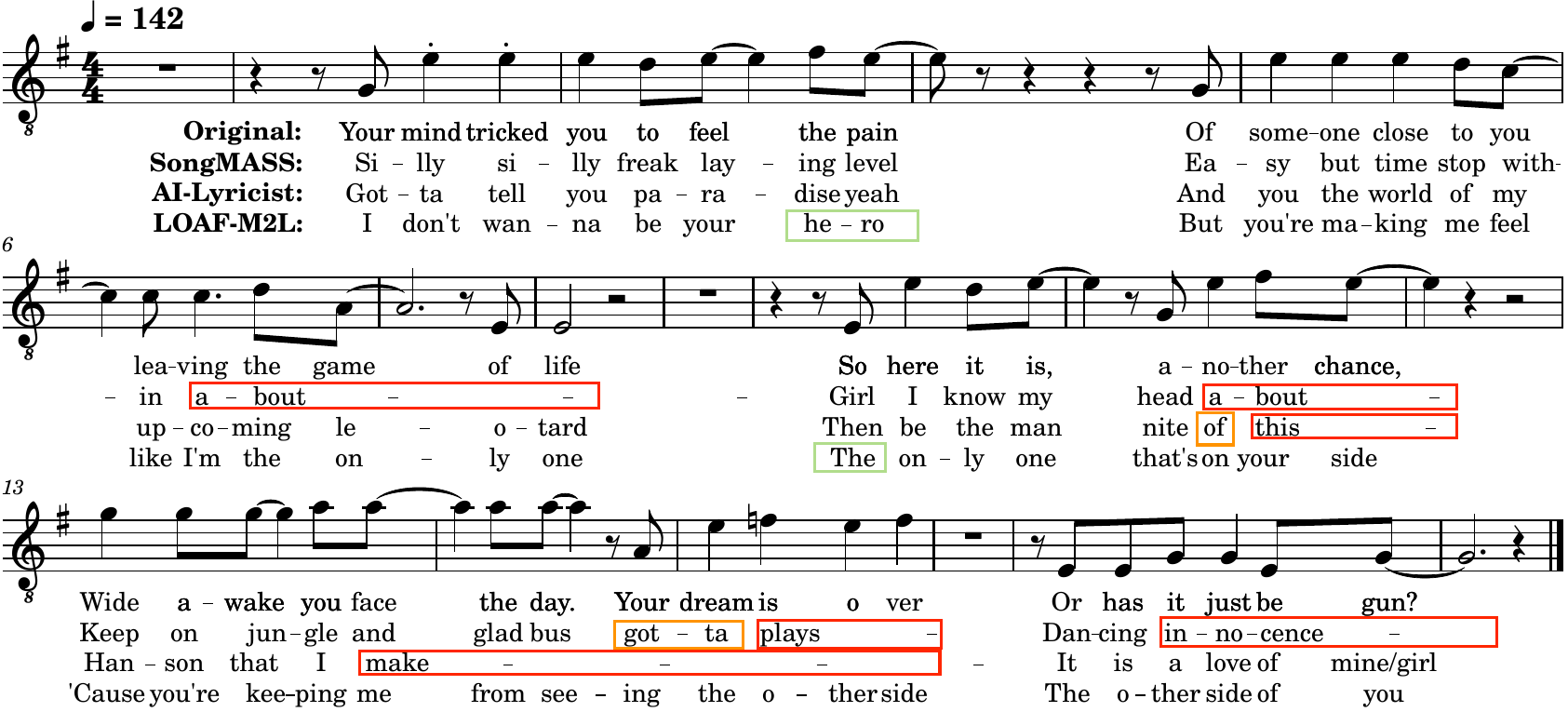}
\caption{\textbf{Caption:} A case of melody with original lyrics, and generated lyrics from different models. \\\hspace{\textwidth} \textbf{Figure 6. Alt Text}: The top row displays the original lyrics, while rows 2 and 3 contain versions by SongMASS and AI-Lyricist, exhibiting issues with length adherence and nuanced compatibility that compromise singability. In contrast, the final row highlights our system's output, which successfully alleviate these issues, offering lyrics with enhanced singability.}
\label{fig:case}
\end{center}
\end{figure}

Figure~\ref{fig:case} presents a qualitative comparison of lyrics generated by different systems for one melody from the subjective evaluation set.

SongMASS produces lyrics that are largely incoherent and often nonsensical. AI-Lyricist generates grammatically complete sentences, but the lines lack semantic continuity because each sentence is produced independently. In contrast, our model generates sentences that are both grammatically sound and semantically coherent across the paragraph.

From a structural perspective, our model achieves perfect one-to-one alignment between notes and syllables, but both SongMASS and AI-Lyricist exhibit multiple mismatches between notes and syllables, as indicated by the red boxes in Figure~\ref{fig:case}.

Our model also shows improvements in fine-grained compatibility. For example (green boxes), the stressed syllable of the word ``hero'' aligns with a melodic peak—an intuitive placement for emphasis. Similarly, the stopword ``the'' in the third sentence is aligned with a short note, matching its low prosodic prominence. In comparison (orange boxes), SongMASS assigns a longer note to the second syllable of ``gotta'' than to the first syllable, resulting in unnatural pronunciation when sung, while AI-Lyricist places the unstressed word ``of'' on a quarter note, producing an awkward prosodic realization.

We host all lyric outputs and some synthesized singing samples used in the human evaluation on an anonymous web page.\footnote{\url{https://5t3dj1lt.github.io/TC0WFOxq}} Readers may refer to this site for other examples and cross-model comparisons.

\section{Conclusion}

This work presents a method for improving the compatibility between generated lyrics and their corresponding melodies. We combine prompt-based length control with a two-stage training framework that integrates self-supervised length-aware learning and supervised melody–lyric fine-tuning. To encourage finer prosodic alignment, we further introduce auxiliary objectives derived from quantifiable music–lyric relationships. 

Both objective metrics and human evaluation show that the proposed system produces lyrics with stronger structural reliability and more consistent melody–lyric compatibility than existing baselines. While the improvements in fine-grained prosody modeling remain modest, the overall gains in singability and alignment are clear.

The proposed framework can support a variety of music-oriented applications, such as assisting songwriters, enhancing music education tools, and enabling personalized or interactive lyric generation. Future work may explore larger models, multilingual extensions, and direct incorporation of expressive singing features.

\section*{Acknowledgements}

This project was funded by research grant A-0008150-00-00 from the Ministry of Education, Singapore.

We used ChatGPT (OpenAI) exclusively for language polishing and improving grammatical clarity during manuscript preparation. The tool was employed to enhance readability and ensure consistent academic style. It was not involved in forming ideas, designing methods, conducting experiments, or interpreting results. All scientific contributions and substantive writing were carried out by the authors.

\bibliographystyle{apacite}
\bibliography{main}

\appendix

\section{Vowel Length Definition}
\label{app:vowel_len}

To approximate the relative duration tendencies of English vowels, we categorize CMU phonemes into three coarse groups based on their typical behavior in singing: short vowels (assigned value 0), long vowels (value 1), and diphthongs (value 2). Specifically, we define the mapping as AH, UH, IH, ER, EH, and AE → 0; AA, IY, UW, and AO → 1; and AY, OW, EY, AW, and OY → 2. While this categorization does not aim to capture the full phonetic complexity of English vowels, it serves as a lightweight heuristic that reflects broad differences in vowel sustain. In singing practice, diphthongs generally require longer or more flexible melodic spans, tense vowels can be sustained more comfortably, and lax vowels tend to occupy shorter durations. Our goal is therefore not precise phonological modeling, but to provide the system with a simple prosodic prior that improves melody--syllable alignment. 

% This heuristic proved effective in practice and avoids the overhead of more elaborate phonetic analysis.

\section{Prompt for ChatGPT}

When adopting ChatGPT as baseline, here is the instruction prompt.
\begin{verbatim}
You are a melody-conditioned lyric generator. I will provide a melody
paragraph composed of several lines. Each line contains a sequence of
notes written as "onset,offset,pitch". Your task is:

1. Treat each melody paragraph independently. Do NOT maintain coherence
   across paragraphs.
2. For each paragraph, generate exactly one lyric line per melody line.
3. Assume one note = one syllable. Your lyrics must match this by
   producing a natural English lyric with the same number of syllables as
   the number of notes in that line.
4. Style is free: poetic, emotional, narrative, atmospheric—anything
   that fits musically.
5. Do NOT include explanations or analysis.
6. Output must be in a single code block.
7. Concatenate lyric lines using the literal string "\n".
8. Text only inside the code block: no hyphens, no extra formatting,
   no syllable markers.

Example input:

"0": [
  "0.95,1.43,45 1.43,1.91,49 ...",
  "5.25,5.73,45 5.73,6.69,57 ...",
  ...
]

You must reply with a code block containing one lyric line per melody
line, joined using "\n".
\end{verbatim}

\section{Ethics Statement}

In our human evaluation, we gathered evaluation scores without personal identifiers to ensure objective and fair comparison. Participants only provided ratings, with no other information being collected. Participation was entirely voluntary, with formal consent obtained from each participant. After participation, evaluators were compensated based on the time they spent completing the questionnaire. We have ensured the questionnaire is free from any offensive content. The process of collecting human annotations has received a review exemption from the Institutional Review Board of the National University of Singapore (NUS-IRB), under Reference Code Number: 2022-813.

\section{Limitations}

Although our system makes measurable progress toward melody-to-lyric generation, the task remains intrinsically challenging, and the model does not produce flawless outputs. As illustrated in the last row of Figure~\ref{fig:intro}, the melody may occasionally place emphasis on a syllable that conflicts with natural pronunciation (e.g., stressing the second syllable of the word ``baby''), resulting in unnatural delivery. Moreover, this work does not fully address several important challenges. One such limitation is the rigidity of length control. While lyrics are influenced by melodic structure, human lyricists do not always adhere strictly to the note-by-note syllable count, and slight deviations are often necessary to preserve semantic clarity or fluency. Introducing controlled flexibility through prompt-based mechanisms is feasible, but achieving stable melody--lyric alignment under flexible length settings remains non-trivial and is left for future exploration.

A second limitation lies in our handling of melody input. The current normalization approach---scaling all timing information by the shortest note duration within each paragraph---may be unstable, as different paragraphs can have different minimum note durations. It may also fail to fully account for variations caused by different time signatures or expressive timing. Future work may benefit from assuming access to musically meaningful metric information such as quantized MIDI with bpm and measure-level structure. Such representations could lead to more robust modeling and improved interpretability of melody-related cues.

\end{document}